\documentclass{article}

\usepackage{aaai21}

\usepackage[utf8]{inputenc} % allow utf-8 input
\usepackage[T1]{fontenc}    % use 8-bit T1 fonts
\usepackage{hyperref}       % hyperlinks
\usepackage{url}            % simple URL typesetting
\usepackage{booktabs}       % professional-quality tables
\usepackage{amsfonts}       % blackboard math symbols
\usepackage{nicefrac}       % compact symbols for 1/2, etc.
\usepackage{microtype}      % microtypography

\usepackage[ruled,vlined]{algorithm2e}
\usepackage{algpseudocode} % algorithm
\usepackage{caption}        % captionof
\usepackage{array}          % thick column hline
\usepackage{booktabs}       % table style
\usepackage{pbox}           % table line break
\usepackage{subcaption}     % multiple figures
\usepackage{graphicx}
\usepackage{mathfixs}
\usepackage{amsfonts}
\usepackage{color}
\usepackage{arydshln}
\usepackage{multirow}
\usepackage{amssymb}

\usepackage{amsmath}
\usepackage{float}

\algnewcommand\algorithmicforeach{\textbf{for each}}
\algdef{S}[FOR]{ForEach}[1]{\algorithmicforeach\ #1\ \algorithmicdo}

\usepackage{amsmath}

\DeclareMathOperator*{\argmin}{arg\,min}

\title{Pair-view Unsupervised Graph Representation Learning}

\author{
}
\author{
  Li You, Luo Binli, Gui Ning\\
    School of Computer Science and Engineering \\
  Central South University \\
  Changsha, China 410083 \\
  \texttt{youli.syvail\@gmail.com; ninggui\@csu.edu.cn} \\
}

\begin{document}
\maketitle
\begin{abstract}

Low-dimension graph embeddings have proved extremely useful in various downstream tasks in large graphs, e.g., link-related content recommendation and node classification tasks, etc. Most existing embedding approaches take node as the basic unit for information aggregation, e.g., node perception fields in GNN or con-textual node in random walks. The main drawback raised by such node-view is its lack of support for expressing the compound relationships between nodes, which results in the loss of a certain degree of graph information during embedding. To this end, this paper pro-poses PairE(\underline{Pair} \underline{E}mbedding), a solution to use “pair”, a higher level unit than a “node” as the core for graph embeddings. Accordingly, a multi-self-supervised auto-encoder is designed to fulfill two pretext tasks, to reconstruct the feature distribution for respective pairs and their surrounding context. PairE has three major advantages: 1) Informative, embedding beyond node-view are capable to preserve richer information of the graph ; 2) Simple, the solutions provided by PairE are time-saving, storage-efficient, and require the fewer hyper-parameters; 3) High adaptability, with the introduced translator operator to map pair embeddings to the node embeddings, PairE can be effectively used in both the link-based and the node-based graph analysis. Experiment results show that PairE consistently outperforms the state of baselines in all four downstream tasks, especially with significant edges in the link-prediction and multi-label node classification tasks.

\end{abstract}
\section{Introduction}
   Graph-structured data are becoming ubiquitous across a broad spectrum of real-world applications, e.g., the networks of molecules, social, biology, finance \cite{Zhang18-survey} and etc. Low-dimensional vector embeddings of nodes in large graphs are extremely useful for a wide variety of tasks such as prediction and graph analysis. Substantial efforts have been committed to developing novel graph embedding approaches. The idea behind graph embeddings usually employs dimensionality reduction techniques to distill high-dimensional information about one node’s neighborhood into a dense vector embedding so that the nodes that are ``close'' in the input graph are also ``close'' in representation space\cite{Hamilton17_review}.In order to achieve high parallelizability and low computation complexity, nodes are normally assumed to be independent of each other and the complex relationships between nodes are largely ignored\cite{Cui18}. By this assumption, the nodes are used as the independent cores for the information condense. The complex information related to a node, e.g. node features, relations, and local structures can be independently processed and condensed into the node embeddings. However, during this embedding process, some complex yet important relation information between nodes is inevitably lost. The analysis tasks on edges, such as link prediction, cannot effectively recover edge information with indirect node embeddings, and it is difficult to obtain the best performance.\cite{wang2020edge2vec}. 

\begin{figure*}
  \centering
  \centerline{\includegraphics[width=.7\textwidth]{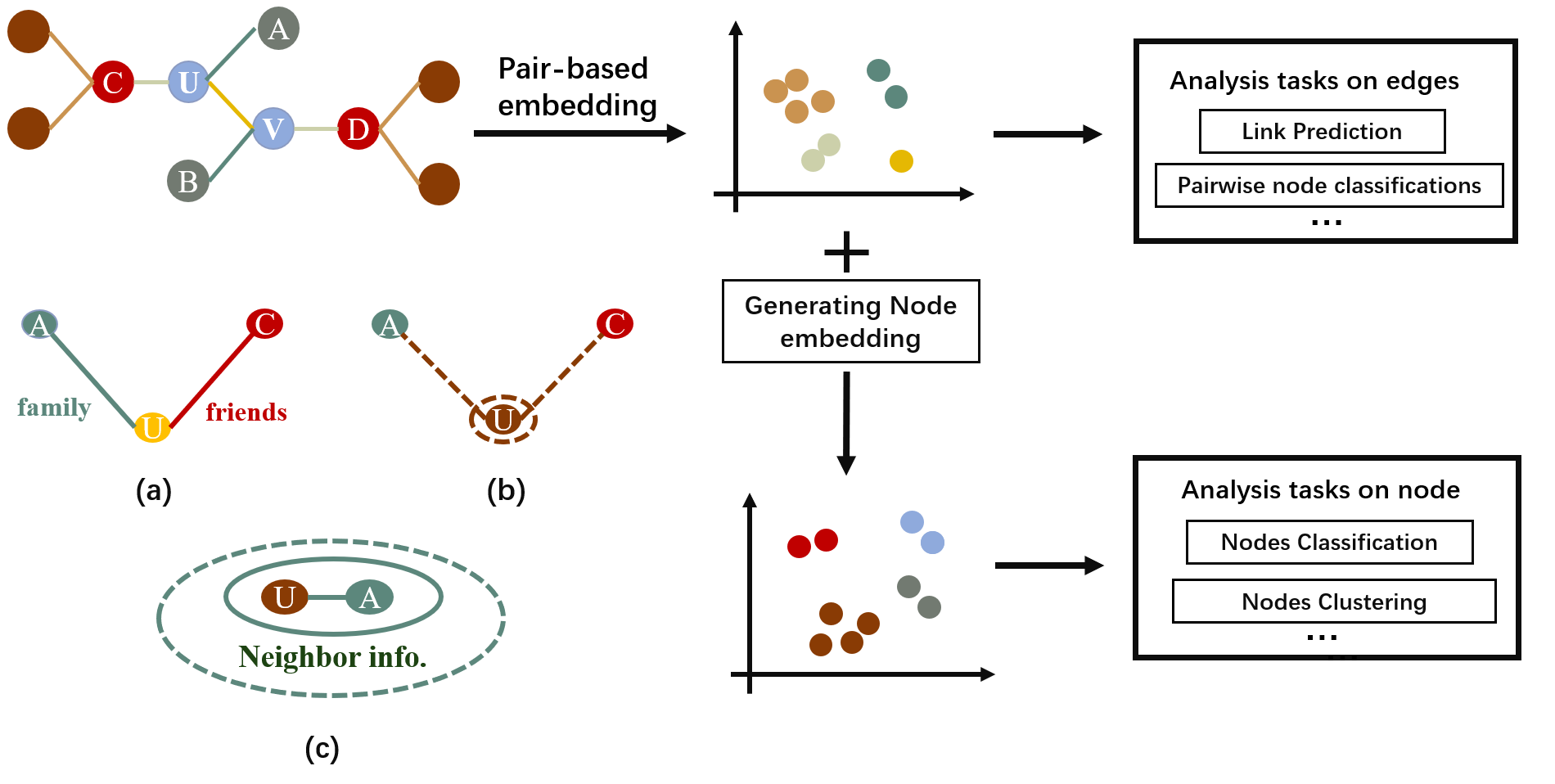}}
   \vspace{-0.1in}
   \caption{a) Node with multiple relationships(roles) b) node-view condense results of the loss of information  c) pair-view with explicit relation representation  d) Pair-view embeddings can be directly used in edge related downstream tasks and translated to node embeddings when dealing with analysis tasks on edges.  
   %It can also be translated into node embeddings for the node analysis
   }
   \vspace{-0.2in}
  \label{fig:view}
\end{figure*}

   Fig.~\ref{fig:view} shows an example of the fact described above.  As shown in Fig.1 (b), the node U have two different relations corresponding to two connections with nodes A and C: family and friends. Conventionally, the two peculiar relations are aggregated by a unified function. The expression of unique characteristics of different types of relationships is limited by the permutation-invariant feature of aggregation operations. In other words, using a single-node view for modeling and aggregating peripheral information into a single node loses its multi-role information and  multi-type relationship. Reference \cite{wang2020edge2vec} also points out that ``representing edges in a direct approach is very important because an edge representation vector generated from the embedding vectors of its endpoints cannot preserve the complete properties of this edge''.

   %\textbf{Thus, the analysis tasks on edges, such as link prediction, cannot achieve the best performance based on these indirect approaches}
   
   To preserve the relation-specific information during embedding, it is argued that we should go beyond the localized node view and utilize higher-level entities with richer expression powers. In this paper, the ``pair'', a structure naturally existing in graphs, is adopted as the basic unit for embedding. As shown in Fig.1 (b),  compared to the node, a unit of ``pair'' has much richer information: the information two paired nodes, the relations between the paired nodes, and the surrounding structural context. Meanwhile, this method has high parallelism.

 %  The losses of information are avoided in the pair-view thanks to the preserved relations between nodes as the pair-view explicitly represents relations

% \begin{figure}[tbh]
% \centering
% % \begin{subfigure}{.33\columnwidth}
% % \centering
% % \includegraphics[width=1.in]{pic/r1.png}
% %  \caption{Node with multiple relationship}
% % \end{subfigure}
% \begin{subfigure}{.35\columnwidth}
% \centering
% \includegraphics[width=1in,height=1.4in]{pic/fig1a.png}
% \caption{Node view}
% \end{subfigure}
% \begin{subfigure}{.6\columnwidth}
% \centering
% \includegraphics[width=1.5in,height=1.5in]{pic/fig1b.png}
% \caption{Pair view}
% \end{subfigure}
% \centering
% \caption{The comparisons of node-view and pair-view }
% %\caption{The comparisons of the node-view and pair-view on nodes \textcolor{red}{with multiple types of relations even in homogeneous networks: the losses of info. in the node-view are avoided in the pair-view thanks to the preserved relations between nodes as the pair-view explicitly represents relations}}
% % \vspace{-0.3cm}
% \label{fig:view}
% \end{figure}

The main contributions are summarized as follows:
 \begin{itemize}
 \item Breaking the shackles of node view, we innovatively propose the use of pair as the basic modeling unit with a so-called ``PairE'' solution. Rather than modeling from a node perspective with little internal structure, PairE explicitly embedding the rich information in, between, and around its paired nodes. The operations related to pair embedding, e.g., data aggregation, pair-node embedding mapping and etc, are correspondingly defined.
%\item Breaking the shackles of node view, PairE innovatively model pair as the basic view unit, and thus is able to explicitly model node relations in embeddings. In the novel solution of pair embedding, more information between nodes, therefore, are preserved. The corresponding operations related to pair embedding, e.g., data aggregation and pair-node embedding mapping, are correspondingly defined. 
 
  \item A multi-task self-supervised auto-encoder is designed to encode information for pairs, which fulfills two mutual-related pretext tasks: to reconstruct both pair features and aggregated features from neighboring nodes. This design allows for the seamless integration of nodes, intra-relations, and structural information into one set of pair embeddings. This design is simple, fast and it demands few hyper-parameters.
 %\item A novel multi-self-supervised auto-encoder is designed to encode the rich information for pairs, which fulfills two mutual-related pretext tasks: one is to reconstruct self-supervised pair features, and another one is to reconstruct aggregated features from neighboring nodes. This multi-task learning allows for the seamless mapping of nodes, intra-relations, and structural information into embeddings. This design is simple and demands fewer hyper-parameters.  
  
\item Experiment results show that PairE outperforms strong baselines in both the link-related and node-related tasks, especially with significant edges in the link-prediction and multi-label node classification tasks.  
\end{itemize}

It is worth noting that PairE differs significantly with those approaches,e.g. \cite{beyond2019} that use existing node-based embedding method to obtain the embedding vectors of edges by converting original graph into the so-called line graph\cite{linegraph}, representing the adjacencies between edges the graph. However, In most graphs, edges normally contains little information and those approaches generally fails to use rich information in nodes. Reference codes and the introduced evaluation data-sets are attached in the Appendix for evaluation.

\section{Related Work}

DeepWalk\cite{Perozzi14} constructs a semantic sequence to extract node representations through the depth-first walk. To integrate more structural information, node2vec\cite{Grover16} allows for the trading off between walks that are more akin to breadth-first or depth-first search and learning embeddings that emphasize community structures and local structure. Line \cite{Tang15} further considers two-hop adjacency neighborhoods. Their approaches did not take the advantage of rich side information, e.g., node features. Regarding the embedding with node features, TADW\cite{Yang15} further incorporates text features of vertices into network representation learning. SINE\cite{Liu18} learns the node embedding by enforcing the alignment of the locally linear relationship between nodes and its K-nearest neighbors on topology and attribute space. DANE\cite{Zhang19_dane} proposes the usage of multiple graph networks to achieve transfer learning. LQANR\cite{Yang19} adopts the usage of K-hop adjacent matrix to jointly learn the low-bit node representations and the layer aggregation weights. To represent the structural information, SDNE\cite{Wang16} uses the auto-encoder to extract both first-order and second-order proximity. Those random walk-based approaches are from node view and normally simplify the complicated relationship between nodes into the co-occurrence probabilities, and can hardly represent complex relations. 

GNN-based solutions typically adopt the so-called message mechanism to collect neighboring information with general differences on how to design the function of aggregation, combine, and readout functions\cite{Xu18_gin}. GCN \cite{Kipf16} uses a weighted sum and a (weighted) element-wise mean. GraphSAGE\cite{hamilton_grapshsage} uses concatenation for combination and adopts the element-wise mean, a max-pooling neural network, and LSTMs aggregators. GAT \cite{Veli17_GAT} proposes the use of self-attention to evaluate the different influences of neighboring nodes. GIN\cite{Xu18_gin} analyzes the expressive power of GNNs and proposes to use the sum aggregator to solve the graph isomorphism test. DGI\cite{Veli18_dgi} relies on maximizing mutual information between patch representations and corresponding high-level summaries of graphs. Besides, in order to include more complex relationships with neighbors, GraphAir\cite{Hu19} proposes the non-linear aggregation of the neighborhood interaction. However, those solutions are still from a single node view and difficult to model the mutual influence between nodes. In other words, these practices decouple the interactive relationship of each node into independent and distributed node representations, and still take node as the basic element for information aggregation and might result in the loss of information during aggregation. 

Few recent researchers noticed the mutual influence between nodes. Splitter\cite{Epasto19} notices the multiple roles played by a node and proposes the usage of multiple representations of the nodes in a graph. However, they still use the node as the basic embedding unit. One recent approach, Edge2vec\cite{wang2020edge2vec} proposes the explicit use of the edge as basic embeddings units. However, this approach considers only the structure information and provides no support on how to translate edge embeddings to node-related tasks.   

\begin{figure*}[t]
\centering
\includegraphics[width=.9\textwidth]{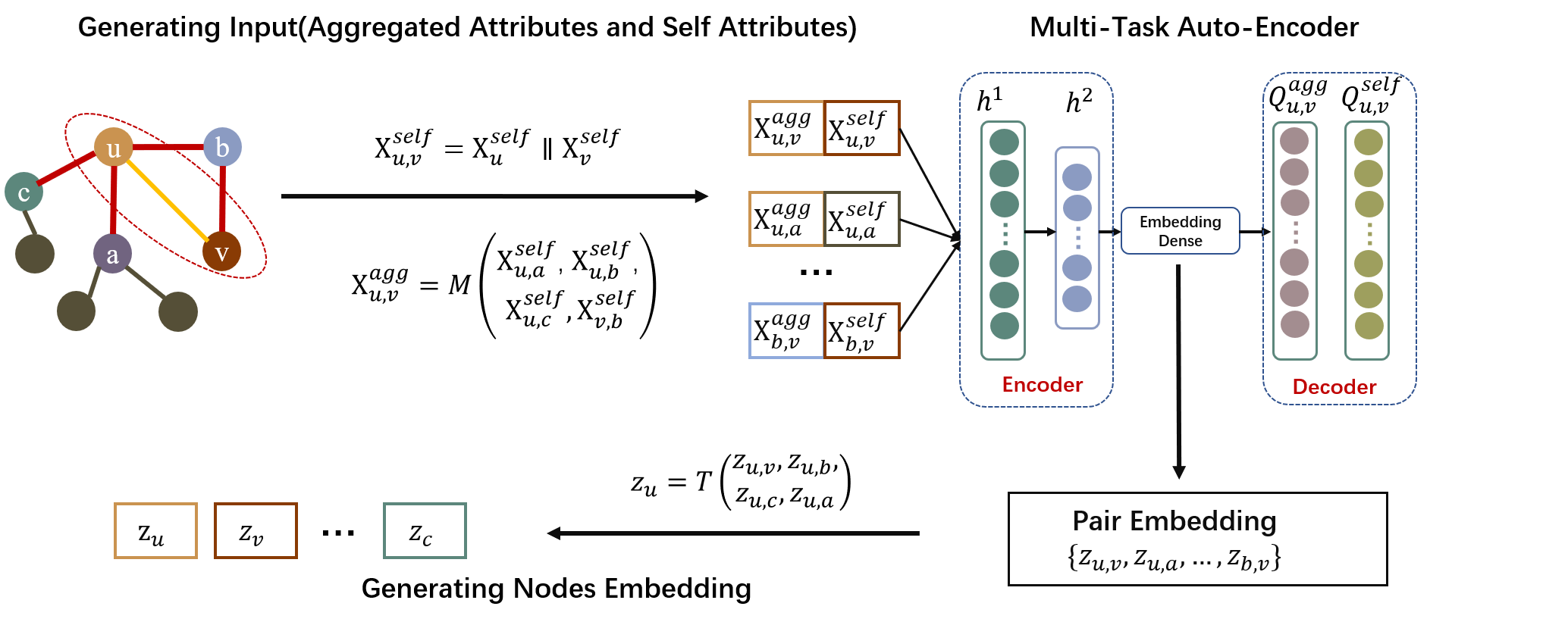}
\caption{A conceptual diagram for multi-self-supervised auto-encoder}
\label{fig:diagram}
 \vspace{-0.2in}
\end{figure*}

% To the best of our knowledge, no previous work has proposed to use the pairs as basic embedding units in the graph representation learning.

% \textbf{Self-supervised learning} aims to learn representations from the data itself without explicit manual supervision. Self-supervised learning has good scalability for data sets of different categories and sizes because it requires no manual labels. In the domain of un-supervised network representation learning, most existing approaches can be generally achieved with a certain pretext task. For instance,One way is context-based, the random-walk-based solution\cite{Perozzi14,Grover16} usually constructs different training targets based on the path (context relationship) of random walks, either via predicting the surrounding nodes through the central node(Skip-gram), or via predicting the central node through the surrounding nodes(CBOW). SDNE\cite{Wang16} and VGAE\cite{Kipf16} reconstruct adjacency matrix or feature matrix by autoencoder. Another Contrastive Based way is to learn the distinctiveness of the nodes. DGI\cite{Veli18_dgi} introduces ``mutual information", taking positive and negative samples in the distribution of input and output, and using Jensen-Shannon divergence to maximize the mutual information of input and output. 

% However,at present, the self-supervised method of the algorithm only limits to a single task. For the graph embedding, it is typical to have information from multiple perspectives to be considered. 
\section{PairE Methods}
\subsection{Preliminaries}

Let $G = (V;E)$ denote a directed graph, where $V$ is the node set and $E$ is the edge set, with node feature vectors $X_v$ for $v \in V$. In order to better differentiate node attributes and aggregated attributes, we generally use $X_v^{self}$ to denote $X_v$.

\noindent \textbf{Definition 1}: (Pair). A pair is an ordered node pair $p=(u,v)$ starting from the node u to the node v, where $u, v \in V$ and $(u,v) \in E$. Node u and v are called the source node and the target nodes, respectively. We denote P(G) as the set of all pairs in G.

\noindent \textbf{Definition 2}: (Pair neighborhood) we define the neighborhood of pair p=(u,v) as a set of pairs adjacent to pair p
    $$N(p_{u,v})= \{n \in P(G) \; | \ |p \cap n| =1\} $$
    
 $N(p_{u,v})$ represents the pair set that has and has only one common node with $p_{u,v}$,including all pairs with u, v as the source node or target node.
  %This neighborhood set $N(p_{u,v})$ defines all pairs that has one and only one nodes in common, for pair $p_{u,v}$, it consists of all the pairs that end with node u or start from node v.
  
\noindent \textbf{Definition 3}: (Pair embedding problem)  The task of pair-based graph embedding is to learn a low-dimensional vector $z_{u,v}$ to better represent the original ordered node pair $p_{u,v}$ in the entire graph. In the embedding process, two ordered nodes and their relationships and surrounding context $N(p_{u,v})$ should be preserved as much as possible.
%The task of pair-based graph embedding is to learn a low-dimensional vector $z_{u,v}$ that can best represent the original ordered node pair $p_{u,v}$ in the entire graph. The ordered two nodes, their relations and and its surrounding context $N(p_{u,v})$ should be preserved as much as possible during the embedding process. 

\noindent \textbf{Definition 4}: (Embedding Translation) To be compatible with different downstream tasks, we define an appropriate translation function $T$ to map the embedding of pairs to a different level of graph elements. In this paper, we only focus on the translation from pair embedding to nodes. We defined a $T(Z_{u,v}) \, \rightarrow \, z_u $,$Z_{u,v} = \{z_{u,v} | \ (u,v) \in N(p_{u,v}) \}$ .
%where $Z_{u,v}$ is the set of pairs' embedding that depart from $u$. 
Due to the order free nature of edges, this function has to be permutation-invariant.

\subsection{Multi-self-supervised auto-encoder}

For \textbf{PairE}, the goal is to find an embedding in which the ``similar'' pairs in the input graph are also ``similar'' in representation space. It is essential that the learned embedding can best represent three types of information: the nodes' self-features, the relations between two paired nodes and the structural information around the pair. 

To learn graph representations from multiple perspectives, a novel multi-self-supervised auto-encoder is designed to simultaneously learn a joint embedding space with data from multiple perspectives. By this auto-encoder, two self-supervised tasks are fulfilled: 1) starting from the pair itself, to fit the features of paired nodes. 2) starting from the neighboring context of the pair, to fit the features of surrounding nodes. These two tasks are inter-correlated. Usually, learning these tasks jointly could improve performance compared to learning them individually~\cite{Zhang17}. Fig.~\ref{fig:diagram} describes the principal structure of the auto-encoder. 

% \begin{algorithm}
% \caption{Algorithm for the multi-self-supervised autoencoder.}
% \label{alg:paire}
% \KwData {Graph $\mathbf{G}=\left(V;{E;X}\right)$}
% %\KwResult{self and agg features$\{x_v^\prime,\ {x_{ps}}_i^\prime,\ \forall v\in\mathcal{V}$, $\forall ps_i\in PS\}$, PairEmbedding}
% \begin{algorithmic}[1]
% %\State {$h_v^0, h_{ps_i} \gets x_v, x_{ps}_i \forall v\in\mathcal{V}, \forall p_{u,v} \in E$ }
% \State $X_{u}^{agg} \gets AGG(X_v) \; \forall p^{u,v} \in E$
% \State $X_{u,v}^{self}=X_{u}^{self} \| X_{v}^{self} \; \forall p^{u,v} \in E$
% \State $X_{u,v}^{agg}= X_u^{agg} \| X_{v}^{agg} \; \forall p^{u,v} \in E$ 
% \state ${z}^1_{self}=\left(W^1*X_{u,v}^{self}}}+b^1\right) $ 

% % \state $X_{u}^{agg} = AGGREGATE({X_{v},\forall v \in N(u)})$
% %     \state $X_{u,v}^{self}=X_{u}^{self} \| X_{v}^{self}$
% %     \state $X_{u,v}^{agg}=X_{u}^{self} \| X_{v}^{self}$
% %    \State{$h_{ps_i}^k\gets \beta \left(W_{ps_i}^k,h_{ps_i}^{k-1}\right)$}
% }
% %\EndFor

% \State {$ h^0={concat(h}_v^2,\ {h_{ps}}_i^2)$} \
%      \For{$k \in K$}
%         \State{$h^{k-1}\gets BatchNormalization(h^{k-1})$}
%         \State{$h^k\gets\mathcal{O}\left(W^k,h^{k-1}\right))$}
%     \EndFor  
%       \For{$k \gets 1$ to $N$}                    
%         \State {$Sum$ $\gets$ {$Sum + A_{k}$}}
%     \EndFor
% \end{algorithmic}
% \label{alg:1}
% \end{algorithm}

\noindent\textbf{Data preparation:} As our pretexts are pair-based, the pair-based data is to be prepared for processing. The pair-specific features are obtained by the direct concatenation of feature vector of nodes $u$ and $v$,  ${X_{u,v}^{self}=X_{u}^{self}} \| X_{v}^{self}$. For the neighboring information, similar to the GNN-based solutions, an aggregation function, {$M\left(p_{u,v}\right)$, is defined to aggregate the local neighborhood attribute information of $p_{u,v}$. There are multiple options for aggregators}. In the current solution, $X_{u,v}^{agg} = X_{u}^{agg} \| X_{v}^{agg}$, which means pair aggregation is the concatenation of the aggregated representation based on paired nodes. Detailed discussion on aggregators can be found in Sec.~3.3.

%, where $\|$ represents concatenation

Thus, the inputs for the autoencoder is $\{X_{u,v}^{self},X_{u,v}^{agg}\} \ \forall p_{u,v}\in E$.
%$\{X_{u}^{self}\|X_{v}^{self},X_{u}^{agg}\|X_{v}^{agg}\} \ \forall (u,v) \in V$.

\noindent\textbf{Encoder:} The encoder is on the right part of Fig.~\ref{fig:diagram}. The inputs of the encoder are  $X_{u,v}^{self}$ and $X_{u,v}^{agg}$ generated from the data preparation process. Firstly, these two types of features pass the two dense layers individually for processing. %The output of $k^{th}$ is denoted as $z^k$, $z^k=\alpha\left(W^k*a^{k-1}}+b^k\right)$. 
To prevent gradient disappearance and achieve more stable performance, we adopt the ``skip connection'', just like ResNet\cite{He16} which skips one or more layers and concatenates the output from the front layer to the back layer.  % ${z}^1_{self}$, ${z}^1_{agg}$, ${z}^2_{self}$ and ${z}^2_{agg}$ . %with the ${z}^3_{self}$ to ${z}^1_{self}\|{z}^1_{agg}\|^2_{self}\|{z}^2_{agg}\|{z}^3_{self}$.
The output of the two dense layers are concatenated and enter the embedding layer. 

% It is worth to note that in order to preserve the linear relationship between paired self-features and its surrounding structure in the newly projected space, no activation function is used in any of those layers. 
% This design is similar to the concept proposed in "locally linear embedding(LLE)" \cite{Roweis00-lle} and later used in SANE\cite{Liu18}. 

Through enforcing the alignment of local relationships from paired self-features and neighboring aggregated information, PairE learns a joint embedding space from both sides simultaneously.

% \textbf{Decoder:} In the decoder located on the right part of Figure 2, \textcolor{red}{two separated dense layers are used to restore the probability distribution: $Q^{Agg}$ and $Q^{Self}$, by sharing the outputs of the embedding layer.}
% % In order to restore the feature distribution represented by a 0/1-valued feature vector for absence/presence indication, the softmax function are employed to represent the categorical distribution, \textcolor{red}{ i.e. a probability distribution over different possible outcomes,} of both self feature $X_{u,v}^{self}$ and aggregated feature $X_{u,v}^{agg}$. 
% Decoder layer uses softmax activation function, the output represents the probability distribution of $X_{u,v}^{self}$ and $X_{u,v}^{agg}$. Thus, the distribution of reconstructed paired self-feature distribution can be calculated with Eq.~\ref{eq:softmax}. Here, $q_k$ denotes the reconstructed value of the $k^{th}$ feature. The calculation of reconstructed aggregated features is the same. 
% \textbf{Decoder:} The output of the two separated dense layers are $H^{Self}$, $H^{Agg}$, and the softmax activation function is used on the output to reconstructed the probability distribution of $X_{u,v}^{Self}$, $X_{u,v}^{Agg}$. The original probability distribution of the input features is $Q^{Agg}$ and $Q^{Self}$, and the reconstructed probability distribution is $Q^{Agg}^{'}$ and $Q^{Self}^{'}$. The specific calculation of $Q^{Self}^{'}$ is as follows,  and $q_k^{self}^{'}$ denotes k-th feature of $Q^{Self}^{'}$. $Q^{Agg}^{'}$ is calculated in the same way.
\noindent\textbf{Decoder:} The output of the two separated dense layers are $H^{self}$ and $H^{agg}$, and the softmax activation function is used on the output to reconstruct the probability distribution of $X_{u,v}^{self}$ and $X_{u,v}^{agg}$. The original probability distributions of input features are denoted as $P^{agg}$ and $P^{self}$, respectively, and the reconstructed probability distributions are denoted as $Q^{agg}$ and $Q^{self}$. 
%The specific calculation of $Q^{self}$ is shown in Eq.(1).
The probability distribution of the $k$-th pair is calculated by Eq.(1).:
%and $Q_k^{self}$ denotes probability distribution of the $k$-th pair. 
\begin{equation} 
\begin{aligned}
\label{eq:softmax}
\setlength{\abovedisplayskip}{6pt}
\setlength{\belowdisplayskip}{6pt}
 Q_k^{self}(i) =  \exp{\left(H^{self}_k(i)\right)}  
  / & \sum_{j\ \in\ F}\exp{\left(H^{self}_k(j)\right)} \\
  &  k\ \in\ E, i\ \in\ F
\end{aligned}
\end{equation}

% \begin{equation}
% \setlength{\abovedisplayskip}{6pt}
% \setlength{\belowdisplayskip}{6pt}
%  q^{agg}_k = \exp{\left(z^{agg}_k\right)}/\sum_{\forall i \in F}\exp{\left({z^{agg}_i\right)}} , \; \ \forall\ \ k\in\ F
% \end{equation}

Where, $Q_k^{self}(i)$ indicates the $i^{th}$ feature's distribution possibility. It is calculated by applying the standard exponential function to each element  
$H^{self}_k(i)$ and the normalized values are divided by the sum of all these exponents. $Q^{agg}$ is calculated in the same way.

The multi-self-supervised auto-encoder has several advantages: Firstly, the data are processed in pairs, and therefore the relationships between nodes are explicitly learned. Secondly, the multi-task processing is realized by training two learning tasks parallelly with a shared embedding layer, and therefore pair content and local neighboring are preserved simultaneously. Meanwhile, the generalization achieved by using the domain-specific information contained for related tasks in training signals is improved. Thirdly, the two learning tasks are mutual-supported since a pair's self-features and its surrounding features are mutual-correlated. The impacts of the corruption or missing certain features can be partially mitigated.  

%The reconstructions of two types of features are supported by the embeddings which contain information for both pair and its surrounding context.

\subsection{Aggregator}
\label{sec:agg}

As shown in the previous section, the inputs of the proposed model depend on the aggregation operations M that produce the aggregated features from a pair' respective neighboring nodes. For the pair $p_{u,v}$,  the aggregated features are the concatenation of the nodes $u$ and $v$'s aggregated features. For the aggregation of node features, many existing solutions have been proposed, e.g., mean, max, LSTM aggregator proposed in GraphSAGE~\cite{hamilton_grapshsage} or the sum aggregator in GIN~\cite{Xu18_gin}. Here, the mean aggregator is adopted as it displays the most stable performance on almost tested datasets. Features of node $u$'s neighbors are aggregated with the element-wise MEAN of the feature vector, as shown in Eq.(2).

\begin{equation} 
\begin{aligned}
   X^{agg}_{u,v} \gets Concat(& M(X_i,\forall i\in N(u)), \\ 
   &  M(X_j,\forall j\in N(v) ))
\end{aligned}
\end{equation}

Where, $i$, $j$ indicate the neighbors for the paired node $u$ and $v$, respectively. We currently define nodes within a node's 1-hop neighborhood. This operation can also be  performed iteratively, similar to the aggregation operations often used in many GNN solutions.

\subsection{Learning embedding}
% In order to learn useful and predictive representations in a fully unsupervised setting, we apply Kullback–Leibler divergence  \cite{Kullback51} to output representations and tune the weight matrices via Adam\cite{Kingma14-adam}. The goal of two tasks is to minimize the reconstruction error in terms of the KL divergence from both aspects. It is the expectation of the logarithmic difference between the original distribution P and the constructed distribution Q. Here, the goal is to minimize the reconstruction errors of KL divergence. 

In order to make the learned embedding representation to recover the original input features as comprehensively and effectively as possible, the Kullback-Leibler (KL) divergence is used to measure the asymmetry of the difference between the two the original probability distribution $P$ and the reconstructed probability distribution $Q$. Therefore, the goals of the two tasks are to minimize the KL divergence between $P^{self}$ and $Q^{self}$, $P^{agg}$ and $Q^{agg}$, respectively.

\begin{equation}
\begin{aligned}
\argmin & ({\rm KL}_{agg}) = \\
 & Min\sum_{i\ \in\ E}\sum_{j\ \in\ F}{P^{agg}_{i}\left(j\right)\ln(\frac{P^{agg}_{i}(j)}{Q^{agg}_i(j)})}
\end{aligned}
\end{equation}
%x\argmin({\rm KL}_{self})=Min\sum_{i\ \in\ E}{p^{self}\left(i\right){p^{self}(i)}/{q^{self}(i)}}
\begin{equation}
\begin{aligned}
 \argmin  & ({\rm KL}_{self}) =\\
& Min\sum_{i\ \in\ E}\sum_{j\ \in\ F}{P^{self}_{i}\left(j\right)\ln(\frac{P^{self}_{i}(j)}{Q^{self}_i(j)})}  
\end{aligned}
\end{equation}

It is worthy to note that rather than making pairs ``close'' in the topology of the graph to have ``close'' representation in the embedding spaces, our solution encodes the pairs with similar features and surrounding contexts to have ``close'' embeddings. Thus, it is possible for two pairs with similar embeddings are far way in the graph. Even in our solution, only the pairs constructed by 1-hop neighbors are trained, the designed encoder is capable of capturing a certain degree of global similarities. 

% \textcolor{red}{Here, the model aims to joint re-construction of information from multiple views by learning representation that are invariant across the node-pair(self) and structural views (aggregation). }

\subsection{Translator}
  As defined in definition 4, to support node-related downstream tasks, e.g., node classification and node clustering, a permutation-free translator function is needed to translate pair embedding to node embeddings.  Similar to the aggregation function, several operations can be used: min, max, mean, and sum. For instance, if the sum operation is adopted, the embedding of node $u$ is the sum of pair embedding with $u$ as a starting point as shown in Eq~\ref{eq:sum}. Sec.~\ref{sec:results} compares the performances of different types of \textit{translators} regarding node classification tasks.

\begin{equation} 
   z_u={Sum}\left(Z_{u,v}\right)  
   %,  \quad      \forall v \in N(u)
 \label{eq:sum}
\end{equation}

\subsection{Computation complexity}
According to Fig.2, the space-time complexity of PairE mainly relies on the number of pairs $|E|$, and the dimension of learned pair features $d$. In the data preparation section, for a given pair p, we only aggregate the characteristics of the first-order neighbors of nodes u and v. Therefore, the time complexity and the space complexity are expressed as $O(|E|)$, $O(F\cdot|E|)$, respectively. Next, we discuss the complexity of neural network computation, every pair is calculated O(epoch) times, then the total computational complexity here is $O(epoch\cdot|E|)$ and the space complexity corresponding to embedding vectors is $O(d\cdot|E|)$, which the dimension of embedding vector. For the translator, the time complexity is expressed as $O(|E|)$. Thus, PairE has an overall time complexity of $O(|E|+epoch\cdot|E|+|E|)$ and the space complexity of $O(F\cdot|E|+d\cdot|E|)$

%1) Due to aggregating data from neighboring nodes, the space complexity is about $O(d|E|)$, where $|E|$ is the size of links and $d$ is the dimension of node features. The computation complexity is about the number of pairs $O(|E|)$. 

% 3) The inductive feature of PairE implies the space complexity could be further reduced to $O(1)$ as data can be generated during computation. Moreover, PairE has low complexity in terms of time \& space, and its scalability therefore is ensured.

\section{Experiments}
\label{sec:experiment}

The performances of PairE are tested on both link-related tasks and node-related tasks to answer the following questions:
\begin{itemize}
    \item whether PairE, from experimental perspective, can encode important pairwise information in its embeddings ?
    \item whether the pair-based embeddings can be effectively translated for downstream node-related tasks while keeps the multiple roles played by one node?
 %   \item whether PairE can effectively encodes multiple roles played by one node ?
\end{itemize}

% 1) Node classification tasks on four different datasets with different training ratios for classification. 2) Node clustering tasks on the same set of datasets evaluated with two different metrics. 3) To check whether PairE recover certain information with the correlation between self-features and agg-features. 

\subsection{Datasets}
\label{sec:setup}

To investigate the learning ability of PairE towards the complex structures from graphs in different application domains, a variety of the real-world graph datasets from benchmarks are selected, i.e. three wildly used citation datasets Cora, CiteSeer, and  Pubmed\footnote{https://linqs.soe.ucsc.edu/data} with one graph, as well as two biological data set with many sub-graphs: Proteins(short for Proteins\_full)~\cite{KKMMN2016} and PPI(protein-protein interaction). Those datasets are selected for evaluation due to their popularity and the general performance of different methods was well demonstrated in the existing literature. Especially, PPI is a  multi-label graph. Each protein might have multiple roles in the graph. Table~\ref{tab:datasets} shows their basic properties.
%Cora contains 2,708 machine learning papers and 5,429 edges in the network. The papers are from 7 classes, which represent the topic of these papers. Citeseer has 3,327 publications from 6 classes and 4732 edges. Pubmed contains 19,717 articles in the field of bio-medicine, 44338 edges and 3 categories. Proteins\_full contains 43471 nodes, 81044 edges and 3 categories .Sec.~\ref{sec:setup} summarizes the compared baselines and experiment settings.

%To probe the ability of PairE to learn the complex structures from graphs in different application domains, we evaluate on a variety of graph datasets chosen from benchmarks commonly used in different downstream tasks. Four real-world networks are used: three wildly used citation datasets Cora, CiteSeer and Pubmed,  \footnote{https://linqs.soe.ucsc.edu/data} as well as one protein dataset PROTEINS-full\cite{KKMMN2016}. The appendix lists their basic statistics and properties. Sec.~\ref{sec:setup} summarizes the compared baselines and experiment settings.

\begin{table}[htbp]
 \centering
  \caption{Basic features of compared datasets}
  \vspace{-0.1in}
    %  \resizebox{0.47\textwidth}{1.2cm}{
    \begin{tabular}{lrrrr}
    \toprule
    \textbf{Dataset} &  \textbf{Nodes} &  \textbf{Edges} & \textbf{Attrs} & \textbf{Classes} \\
    \midrule
    Cora  & 2708  & 5429  & 1433  & 7 \\
    Citeseer & 3327  & 4732  & 3703  & 6 \\
    Pubmed & 19717  & 88651  &  500    & 3  \\
    Proteins & 43471 & 81044 & 29    & 3 \\
    PPI  &  56944    &    818716    &  50  &  121 \\
    \bottomrule
    \end{tabular}%
  \label{tab:datasets}%
  \vspace{-0.1in}
 % }
\end{table}%

% 1) Node classification tasks on four different datasets with different training ratios for classification. 2) Node clustering tasks on the same set of datasets evaluated with two different metrics. 3) To check whether PairE recover certain information with the correlation between self-features and agg-features. 

%3) Performances when partial attributes are masked to check whether PairE can tolerate a certain degree of information loss. 

\subsection{Baseline models}
In terms of node classification and node clustering, the PairE is compared with several state-of-the-art unsupervised baselines and two semi-supervised baselines:
% \begin{itemize}
 \noindent  \textbf{DeepWalk}\cite{Perozzi14}, the first approach for learning latent representations of vertices in a network with NLP.
 
  \noindent  \textbf{TADW}\cite{Yang15}, a representative factorization-based approach that incorporates node content features via inductive matrix factorization. 
  
  \noindent \textbf{SINE}\cite{Zhang_sine}, a network embedding algorithm with good capability for missing attributes in an incomplete graph; NodeE, is selected for the ablation study. It works on the node level embedding with the same multi-self-supervised auto-encoder as PairE.
  
  \noindent\textbf{ProNE}~\cite{zhang2019prone}: is a network embedding algorithm approach based on matrix factorization without node features. It enhanced the embeddings by propagating them in the spectrally modulated space.
  
  \noindent \textbf{GraphSAGE}~\cite{hamilton_grapshsage}, an inductive framework that leverages node feature information to generate node embeddings for previously unseen data; Here, the mean aggregator is used. 
  
  \noindent \textbf{DGI}~\cite{Veli18_dgi}, a contrastive self-supervised approach via maximizing mutual information between patch representations and corresponding high-level summaries of graphs. A variant of DGI-GCN is used due to its superior performance.

%  \noindent \textbf{P-GNN}~\cite{P-GNN2019}, a position-aware GNN solution, an recent GNN solution with augmented position information. This solution is optimized for link related tasks as its embeddings contains global position information. 
%  \item GCN(graph convolutional network)\cite{gcn2016} a semi-supervised approach to learning over graph structures using convolution operators for embeddings. This method is choose to check whether our performance comparable with semi-supervised solutions.  
 %\end{itemize}
 
The implementation of ProNE, SINE are from Karateclub~\cite{karateclub}\footnote{https://github.com/benedekrozemberczki/karateclub}, and GraphSage, DGI algorithm are from stellargraph~\cite{StellarGraph}\footnote{https://github.com/stellargraph/stellargraph}. These implementations have matching with or even better performance than the original paper. The implementations of DeepWalk and TADW are from their corresponding projects on Github.
%\textcolor{red}{  All the implementation are from their original paper except for TADW are implemented by ourselves. }

% The embedding dimensions of all compared baselines are set to be 128 and the batch size for PairE is set to be 1024. Epoch is set to be 30.

\begin{table*}[ht]
  \begin{center}
  \caption{Averaged ROC AUC values for Link prediction and Pairwise node classification tasks, values before / is for Link prediction and after for Pairwise node classification.}
  \vspace{-0.1in}
%   \resizebox{1.0\textwidth}{14mm}{
    \renewcommand\arraystretch{1.2}
    \begin{tabular}{l|c|c|c|c}
    \hline
    Name  & Cora & Citeseer & Pubmed & Proteins \\
   \hline
    DeepWalk &   .8117/.8057   & \underline{.8193}/.7296      &   .7995/.7201      & \underline{.9310}/.5411       \\
    TADW  & .6591/.7526  & .6925/.7026      &  .7054/.7350      &  .5090/\textbf{.7530}       \\
    SINE  & .7184/.8199  & .7671/\underline{.7367} & .7147/.7274 & .7491/\underline{.6207}  \\
    
    ProNE  & \underline{.8475}/.8090  & .8294/.7334 & .8371/.7626 & .9226/.5544  \\
    
    GraphSAGE & .8286/\underline{.8341} & .7902/.7169 & \underline{.8648}/.7193 & .8469/.6012  \\
    DGI-gcn & .6701/.8261  & .6725/.6804 & .7609/.7468 & .4751/.5788  \\
 %   P-GNN & \underline{.8217}/.7615  & .7791/.6804 & .8276/.6527 & .8224/\underline{.7290} & .7695/--\\
 %  \midrule
 %   GCN & .8764/  & .8888/ & .9013/ & .8718/ \\
 %\midrule
 %   NodeE & /.7173 & ./.6722 & ./.6904 & ./.5859 \\
    PairE & \textbf{.9250}/\textbf{.8607} & \textbf{.9157}/\textbf{.7554} & \textbf{.9499} /\textbf{.7926} & \textbf{.9352}/.6003  \\
    \hline
    \end{tabular}
    % }
  \label{tab:linkpre}%
%    \vspace{-0.2in}
\end{center}
\end{table*}%

\begin{table*}[h]
  \begin{center}
  \caption{Node classification results for compared datasets, evaluated with average Micro-F1. Analogous trends hold for averaged Macro-F1 scores. Values for PPI are for the multi-label node classification task.}
  \label{tab:node}
    \vspace{-0.1in}
  % \resizebox{1.0\textwidth}{1.9cm}{
  \renewcommand\arraystretch{1.1}
    \begin{tabular}{p{4.5em}|ccc|ccc|ccc|ccc}
    \hline
    Name &\multicolumn{3}{c|}{Cora}&\multicolumn{3}{c|}{Citeseer}&\multicolumn{3}{c|}{Pubmed}&\multicolumn{3}{c}{PPI} \\
    %\midrule
    Ratio & 30\% & 50\% & 70\% & 30\% & 50\% & 70\% & 30\% & 50\% & 70\% & 30\% & 50\% & 70\% \\
    \hline
    DeepWalk&.8005&.7939&.8237&.5602&.5809&.5907&.7980&.8016&.7994&.5283&.5375&.5433
    \\
    %&.5204&.5279&.5232\\
    TADW &\underline{.8202}&\underline{.8391}&\underline{.8475}&.7145&.7336&\underline{.7410}&.8155& .8230&.8289&.4780&.4905&.4768\\
    %&.4851&.5184&.5158\\
    ProNE &.7747&.7968&.8137&.5431&.5550&.5730&.7858&.7952&.7970&.5419 &.5501 &.5419\\
    
    SINE &.7674&.7740&.7831&.6263&.6434&.6727&.8546&\underline{.8683}&\underline{.8698}&.5046 &.5194 &.5013\\
    %&.6897&.6453&.6895\\
    \small{GraphSAGE} &.8018&.8361&.8290&\underline{.7296}&\underline{.7337}&.7362&.8382&.8437&.8383&.6697&.6642&.6615\\
    %&\underline{.7534}&\underline{.7657}&\underline{.7732} \\
    DGI-GCN &.8128&.8257&.8337&.6894&.6963&.7126&.8311&.8328&.8329&\underline{.9567}&\underline{.9565}&\underline{.9618}\\
    %&.6414&.6453&.6494  \\
%
% \midrule
 %   NodeE&.8103&.7962&.7979&.7074&.7131&.7311&\underline{.8565}&\.8642&.8660&.7383&.7423&.7476\\
    PairE&\textbf{.8651}&\textbf{.8692}&\textbf{.8728}&\textbf{.7535}&\textbf{.7492}&\textbf{.7496} &\textbf{.8857}&\textbf{.8793}&\textbf{.8828}&\textbf{.9825}&\textbf{.9834}&\textbf{.9828}\\
    %&\textbf{.7595}&\textbf{.7760} &\textbf{.7838}\\
    \hline
    \end{tabular}
    \end{center}
%    }
\end{table*}

\subsection{Experimental setup}

The link-related task including link prediction and pairwise node classification tasks. The link prediction task is a classical problem for the graph to predict whether a given edge which does not exist or will appear in the future. The pairwise node classification task predicts whether a pair of nodes belong to the same community/class and used in \cite{P-GNN2019}. The node-related tasks include both the widely used node classification task and the node clustering task.
 
\noindent\textbf{Inductive learning for link-related tasks} We use a similar prepossessing step in~\cite{Grover16}, which first employs the graph embedding method in the link prediction task. We first remove 20\% edges from the original network randomly while maintaining graph connectivity. The edges in the residual network are treated as positive samples in the training set and the removal edges are treated as positive samples for the test sets. We sample node pairs that are not connected in the residual network as negative training samples, and node pairs that are not connected in the origin network as negative test samples. In the corresponding set, the negative samples are with the same numbers of the positive ones. Note that we do not allow the model to observe ground-truth graphs at the training time. For the pairwise node classification task, we use the setting in P-GNN~\cite{P-GNN2019} to predict whether a pair of nodes belong to the same community/class.
%In this case, a pair of nodes that do not belong to the same community are a negative example. 
Due to the fact that there is lack of well-accepted setting for multi-label pairwise node classification. PPI is not used for link-related tasks.

\noindent\textbf{Transductive setting for node-related tasks} In the node-related tasks, the transductive setting is adopted. During the embedding process, all the nodes are used in the embedding generation processes. The embeddings of vertices from different solutions are taken as the features to train classifiers with different training ratios, from 30\%, 50\% to 70\%, and classification accuracy is evaluated with remaining data. For the node clustering task, node representations learned by different solutions are fed into the K-means clustering algorithm, and the respective categories are grouped with the K-means algorithm. The average accuracy (ACC) and Normalized Mutual Information(NMI) values are used for evaluation. 

\noindent\textbf{Experiment settings}
These embedding methods are all node-based, and thus we need to generate vectors for edges according to the embedding vectors of its source node and the target node. Here, the approaches in Node2Vec~\cite{Grover16} with Hadamard product of the two vectors are not appropriate as it generates the same vector for pair (u, v) and (v, u) given the embedding vectors of u and v. Thus, we concatenate the embedding vectors of u and v to generate the vector representation of pair. The embedding dimension of both pair embeddings in PairE and node embeddings for other baselines are all set to 128. Other parameters are set as the default settings in the corresponding paper. For the PairE, we set epoch 30 and batch size 1024. A classifier with Logistic Regression (LR) algorithm is used with the one-vs-rest strategy. we report results over 10 runs
with different random seeds and train/validation splits and the averaged ROC AUC/Micro-F1 score is reported. We fix model configurations across all the experiments. Details for experiment settings are provided in the Appendix.

\subsection{Results and Analysis}
\label{sec:results}
   
\subsubsection{Link prediction} For the link prediction task, two nodes generally more likely to form a link, if match the neighboring pattern existed in the graph. This neighboring pattern can be whether they are close together or their features have certain matches. Therefore, the task can largely benefit from pair-wise embeddings. Results for the link prediction task are reported in Tab.~\ref{tab:linkpre}. We observe that Pair-E significantly outperforms other strong baselines across all datasets over other baselines. For instance, it improves ROC AUC score by up to 40.3/12.6\% over the best/worst compared baseline in Cora and 36.2/11.7\% in Citeseer and 34.6/11.0\% in Pubmed. These results clearly show that PairE benefits from preserving the pair structure of the graphs. For 
the Proteins with many sub-graphs, The deepwalk based on the random walk strategy is conducive to capturing sub-graph information, so it performs outstandingly on the dataset.

\subsubsection{Pairwise node classifications} For this task, it is important to model the relationship between two connected nodes. Thus solutions that focus on learning node-view structure-aware embeddings will not perform well in these tasks. 

Tab.~\ref{tab:linkpre} summarizes the performance of all the compared solutions. In all the well-studied citation datasets, PairE maintains considerable advantages over other baselines. The only exception is Proteins. This dataset contains many sub-graphs. TADW, based on the matrix factorization techniques, is comparably easy to  catch this sub-graph information.
  
\subsubsection{Node classification} This task is a well-adopted node-based task. However, as shown in Tab.~\ref{tab:node}, even some strong baselines are up-to-date, PairE outperforms by a significant margin in four tested datasets. Moreover, PairE achieves very stably performance within all training ratio ranges across different datasets, even for the well-adopted three citation datasets. Specifically, in the multi-labeled PPI dataset, PairE achieves the averaged F1 score to 0.98 range, up to 100\% relative performance improvement. It shows that PairE can capture multiple roles played by individual nodes. It is very important for many graph analysis tasks. 
%SINE has very good performance in Pubmed while comparable weak performance in Citeseer. TADW has strong performance in the citation datasets while inferior performance in the Protein-full dataset.

Furthermore, PairE shows more significant performance advantages when the training ratio is small. This fact clearly shows that the manifold obtained by PairE is much clearer and much easier to be learned by classifier than other baselines. Thus, the classifier can learn a model with a small portion of the embeddings.

\subsubsection{Node clustering}
  As a complement to node classification, node clustering experiments are conducted to examine the embedding quality from other perspectives. Due to pape limits, only results for Cora and Citeseer are reported in Tab.~\ref{tab:clustering}. It is easy to see that PairE achieves the best performance in terms of both metrics with a considerable edge. Similar performances have been achieved in other datasets. Please refer to the Appendix for details. 
  
  %and group them into respective categories. 
% \vspace{-0.1cm} 
\begin{table}[htbp]
  \centering
  \caption{Node Clustering Results on Cora and Citeseer}
  \vspace{-0.1cm} 
%     \resizebox{.75\textwidth}{1.7cm}{
    \renewcommand\arraystretch{1.1}
    \begin{tabular}{l|cc|cc}
    \hline
    Name & \multicolumn{2}{c|}{Cora} & \multicolumn{2}{c}{Citeseer} \\% & \multicolumn{2}{c|}{Pubmed} & \multicolumn{2}{c}{Proteins\_full} \\
    Metrics & NMI & ACC & NMI & ACC \\
    \hline
    DeepWalk &  .3259  & .4025  &  .1960  & .3309 \\ %& .2536  & .5776  & .0003  & .4069 \\
    TADW & .4759  & .6027  & .3711  & .5987 \\ %& .2575  & .6097  & .0004  & .3893 \\
    SINE & .4858  &  .6861  & .3519  & .6186 \\ %& \textbf{.3158}  & \textbf{.6902}  & .0005  & .3665 \\
    GraphSAGE & .5142 & \underline{.6961} & \underline{.3770} & \underline{.6299} \\ %& .2307 & .6335 & .0010 & .3554 \\
    DGI-GCN & \underline{.5504} & .6913 & .3576 & .6087 \\ %& .2477 & .6476 & .0004 & .3996 \\
  %  NodeE & .4571  & .6680  & .3217  & .5888  \\ %& \underline{.2696}  & .6351  & \underline{.0020}  & \underline{.4716} \\
    PairE & \textbf{.5585} & \textbf{.7149} & \textbf{.4251} & \textbf{.6501} \\%& .2535 & \underline{.6489} & \textbf{.0062} & \textbf{.5913} \\
    \hline
    \end{tabular}%
%    }
  \label{tab:clustering}%
% \vspace{-0.1in}
\end{table}%

\subsection{Comparisons of translators}
\label{sec:translator}
In this section, the performance impacts of different translators in PairE, i.e. sum, mean, max and min, are analyzed and evaluated based on the node classification task. 

\begin{figure}[htb]
\centering
\begin{subfigure}{.50\columnwidth}
\centering
 \includegraphics[width=1.8in,height=1.4in]{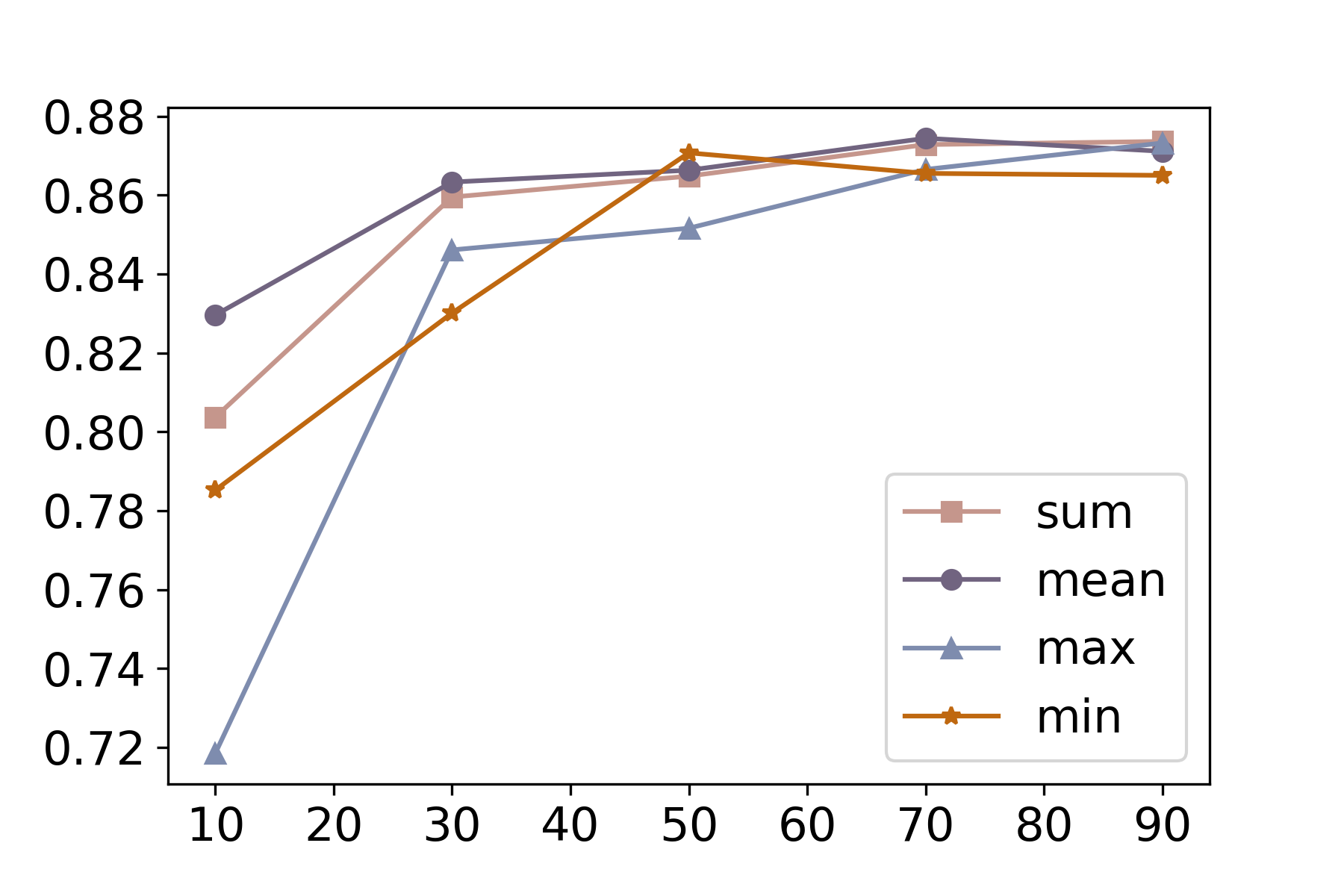}
  \vspace{-0.2cm} 
 \caption{Cora}
\end{subfigure}
\begin{subfigure}{.48\columnwidth}
\centering
\includegraphics[width=1.8in,height=1.38in]{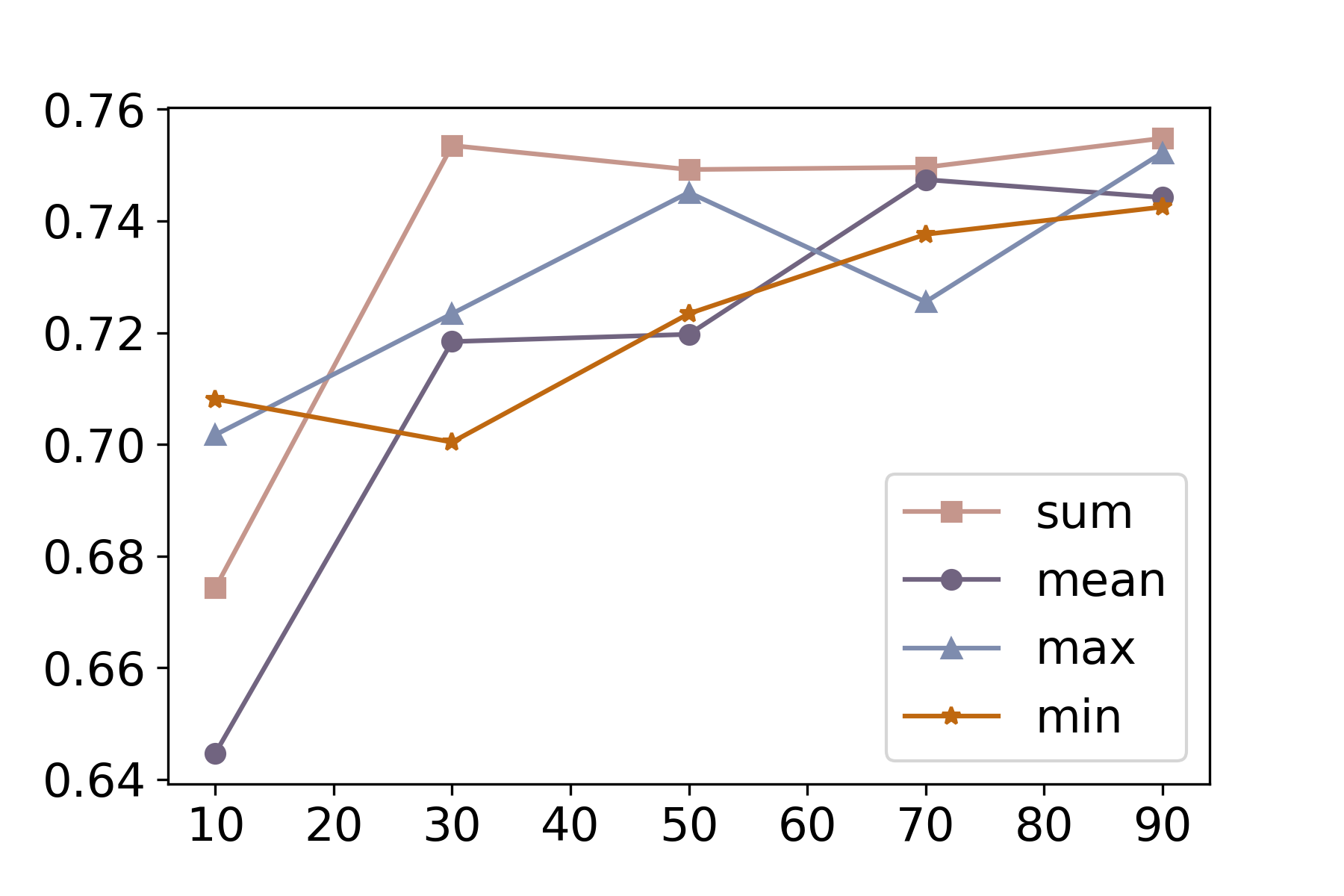}
 \vspace{-0.cm} 
\caption{CiteSeer}
\end{subfigure}
\centering
 \vspace{-0.cm} 
\caption{Translators comparisons on Cora and Citeseer with different trainning ratios }
\vspace{-0.cm}
\label{fig:translator}
\end{figure}

As can be seen from Fig.~\ref{fig:translator}, the sum translator generally achieves the best performance, among the four compared translators. The sum translator performs the embedding accumulation of all pairs starting from the node. To some extent, the ``sum'' translator keeps certain information on the number of connected edges. This peculiar feature might explain its comparably better performance. The other three have comparably unstable performance. For instance, the ``mean'' translator performs well in the Cora dataset while relatively poor in the CiteSeer dataset. However, what operation should be used to best translate pair embeddings to node embeddings with minimal information loss remains a largely unexplored area and could be an interesting direction for future work.

\section{Conclusion}
\label{sec:conclusion}
 The real-world networks usually contain complex relationships between nodes, even in many so-called homogeneous networks. 
In order to preserve that relationship information as much as possible, this paper proposes a novel multi-self-supervised auto-encoder, so-called PairE, that is able to learn graph embeddings from pair-view so as to avoid the information loss inevitably occurred in the node-view embedding approaches. 
%A certain degree of information is inevitably lost due to the node-view in most  existing graph embedding approaches.
%Most existing graph embedding approaches normally take a node-view. Thus, a certain degree of information loss is inevitable. 
%In this paper, we propose a novel multi-self-supervised auto-encoder that can learn graph embeddings from a higher level view, i.e. pair-view. 
The seamless integration of node self-features and nearby structural information is achieved with shared layers in the multi-task auto-encoder design. We also design a systematic approach to translate pair embeddings to node embeddings for node-related graph analysis tasks. 
%The multi-task auto-encoder design with shared layers facilitates the seamless integration of node self feature and nearby structural information.
Four types of downstream tasks, including both link-related and node-related tasks, are tested for validation.
%The resulting solution is tested on both node classification and node clustering tasks. 
Extensive experiments show that PairE outperforms strong baselines in all those tasks with significant edges. 
\clearpage

\bibliographystyle{plain}
\bibliography{graph}

\begin{thebibliography}{10}

\bibitem{beyond2019}
Sambaran Bandyopadhyay, Anirban Biswas, M~Narasimha Murty, and Ramasuri
  Narayanam.
\newblock Beyond node embedding: A direct unsupervised edge representation
  framework for homogeneous networks.
\newblock {\em arXiv preprint arXiv:1912.05140}, 2019.

\bibitem{linegraph}
Jørgen Bang-Jensen and Gregory~Z Gutin.
\newblock {\em Digraphs: theory, algorithms and applications}.
\newblock Springer Science \& Business Media, 2008.

\bibitem{Cui18}
Peng Cui, Xiao Wang, Jian Pei, and Wenwu Zhu.
\newblock A survey on network embedding.
\newblock {\em IEEE Transactions on Knowledge and Data Engineering},
  31(5):833--852, 2018.

\bibitem{StellarGraph}
CSIRO's Data61.
\newblock Stellargraph machine learning library.
\newblock \url{https://github.com/stellargraph/stellargraph}, 2018.

\bibitem{Epasto19}
Alessandro Epasto and Bryan Perozzi.
\newblock Is a single embedding enough? learning node representations that
  capture multiple social contexts.
\newblock In {\em The World Wide Web Conference}, pages 394--404, 2019.

\bibitem{Grover16}
Aditya Grover and Jure Leskovec.
\newblock node2vec: Scalable feature learning for networks.
\newblock In {\em Proceedings of the 22nd ACM SIGKDD international conference
  on Knowledge discovery and data mining}, pages 855--864, 2016.

\bibitem{hamilton_grapshsage}
Will Hamilton, Zhitao Ying, and Jure Leskovec.
\newblock Inductive representation learning on large graphs.
\newblock In {\em Advances in neural information processing systems}, pages
  1024--1034, 2017.

\bibitem{Hamilton17_review}
William~L Hamilton, Rex Ying, and Jure Leskovec.
\newblock Representation learning on graphs: Methods and applications.
\newblock {\em arXiv preprint arXiv:1709.05584}, 2017.

\bibitem{He16}
Kaiming He, Xiangyu Zhang, Shaoqing Ren, and Jian Sun.
\newblock Deep residual learning for image recognition.
\newblock In {\em Proceedings of the IEEE conference on computer vision and
  pattern recognition}, pages 770--778, 2016.

\bibitem{Hu19}
Fenyu Hu, Yanqiao Zhu, Shu Wu, Weiran Huang, Liang Wang, and Tieniu Tan.
\newblock Graphair: Graph representation learning with neighborhood aggregation
  and interaction.
\newblock {\em arXiv preprint arXiv:1911.01731}, 2019.

\bibitem{KKMMN2016}
Kristian Kersting, Nils~M. Kriege, Christopher Morris, Petra Mutzel, and Marion
  Neumann.
\newblock Benchmark data sets for graph kernels, 2016.

\bibitem{Kipf16}
Thomas~N Kipf and Max Welling.
\newblock Semi-supervised classification with graph convolutional networks.
\newblock {\em arXiv preprint arXiv:1609.02907}, 2016.

\bibitem{Liu18}
Weiyi Liu, Zhining Liu, Toyotaro Suzumura, and Guangmin Hu.
\newblock Sane: Scalable attribute-aware network embedding.
\newblock 2018.

\bibitem{Perozzi14}
Bryan Perozzi, Rami Al-Rfou, and Steven Skiena.
\newblock Deepwalk: Online learning of social representations.
\newblock In {\em Proceedings of the 20th ACM SIGKDD international conference
  on Knowledge discovery and data mining}, pages 701--710, 2014.

\bibitem{karateclub}
Benedek Rozemberczki, Oliver Kiss, and Rik Sarkar.
\newblock {Karate Club: An API Oriented Open-source Python Framework for
  Unsupervised Learning on Graphs}.
\newblock In {\em Proceedings of the 29th ACM International Conference on
  Information and Knowledge Management (CIKM '20)}. ACM, 2020.

\bibitem{Tang15}
Jian Tang, Meng Qu, Mingzhe Wang, Ming Zhang, Jun Yan, and Qiaozhu Mei.
\newblock Line: Large-scale information network embedding.
\newblock In {\em Proceedings of the 24th international conference on world
  wide web}, pages 1067--1077, 2015.

\bibitem{Veli17_GAT}
Petar Veličković, Guillem Cucurull, Arantxa Casanova, Adriana Romero, Pietro
  Lio, and Yoshua Bengio.
\newblock Graph attention networks.
\newblock {\em arXiv preprint arXiv:.10903}, 2017.

\bibitem{Veli18_dgi}
Petar Veličković, William Fedus, William~L Hamilton, Pietro Liò, Yoshua
  Bengio, and R~Devon Hjelm.
\newblock Deep graph infomax.
\newblock {\em arXiv preprint arXiv:1809.10341}, 2018.

\bibitem{wang2020edge2vec}
Changping Wang, Chaokun Wang, Zheng Wang, Xiaojun Ye, and Philip~S Yu.
\newblock Edge2vec: Edge-based social network embedding.
\newblock {\em ACM Transactions on Knowledge Discovery from Data (TKDD)},
  14(4):1--24, 2020.

\bibitem{Wang16}
Daixin Wang, Peng Cui, and Wenwu Zhu.
\newblock Structural deep network embedding.
\newblock In {\em Proceedings of the 22nd ACM SIGKDD international conference
  on Knowledge discovery and data mining}, pages 1225--1234, 2016.

\bibitem{Xu18_gin}
Keyulu Xu, Weihua Hu, Jure Leskovec, and Stefanie Jegelka.
\newblock How powerful are graph neural networks?
\newblock {\em arXiv preprint arXiv:1810.00826}, 2018.

\bibitem{Yang15}
Cheng Yang, Zhiyuan Liu, Deli Zhao, Maosong Sun, and Edward Chang.
\newblock Network representation learning with rich text information.
\newblock In {\em Twenty-Fourth International Joint Conference on Artificial
  Intelligence}, 2015.

\bibitem{Yang19}
Hong Yang, Shirui Pan, Ling Chen, Chuan Zhou, and Peng Zhang.
\newblock Low-bit quantization for attributed network representation learning.
\newblock In {\em Twenty-Eighth International Joint Conference on Artificial
  Intelligence {IJCAI-19}}. International Joint Conferences on Artificial
  Intelligence Organization, 2019.

\bibitem{P-GNN2019}
Jiaxuan You, Rex Ying, and Jure Leskovec.
\newblock Position-aware graph neural networks.
\newblock {\em arXiv preprint arXiv:1906.04817}, 2019.

\bibitem{Zhang18-survey}
Daokun Zhang, Jie Yin, Xingquan Zhu, and Chengqi Zhang.
\newblock Network representation learning: A survey.
\newblock {\em IEEE transactions on Big Data}, 2018.

\bibitem{Zhang_sine}
Daokun Zhang, Jie Yin, Xingquan Zhu, and Chengqi Zhang.
\newblock Sine: Scalable incomplete network embedding.
\newblock In {\em 2018 IEEE International Conference on Data Mining (ICDM)},
  pages 737--746. IEEE, 2018.

\bibitem{zhang2019prone}
Jie Zhang, Yuxiao Dong, Yan Wang, Jie Tang, and Ming Ding.
\newblock Prone: fast and scalable network representation learning.
\newblock In {\em Proc. 28th Int. Joint Conf. Artif. Intell., IJCAI}, pages
  4278--4284, 2019.

\bibitem{Zhang19_dane}
Yizhou Zhang, Guojie Song, Lun Du, Shuwen Yang, and Yilun Jin.
\newblock Dane: Domain adaptive network embedding.
\newblock {\em arXiv preprint arXiv:1906.00684}, 2019.

\bibitem{Zhang17}
Yu~Zhang and Qiang Yang.
\newblock A survey on multi-task learning.
\newblock {\em arXiv preprint arXiv:1707.08114}, 2017.

\end{thebibliography}
\end{document}